\documentclass[letterpaper, 10 pt, conference]{ieeeconf}
\IEEEoverridecommandlockouts
\usepackage{graphicx}
\usepackage{stfloats}
\usepackage{times}
\usepackage{float}
\usepackage{epsfig}
\usepackage{amsmath}
\usepackage{bm}
\usepackage{amssymb}
\usepackage{booktabs}
\usepackage{subcaption}
\usepackage{wrapfig}
\usepackage{makecell}
\usepackage{rotating}
\usepackage{multirow}
\usepackage{xcolor}
\usepackage{adjustbox}
\usepackage{cuted}
\usepackage{colortbl}
\usepackage{tcolorbox}
\usepackage{array}

\makeatletter
\let\labelindent\relax
\makeatother
\usepackage{enumitem}

\usepackage{multicol}
\usepackage{pifont}
\usepackage{bigdelim}
\usepackage{setspace}
\usepackage{blindtext}
\usepackage{url}
\usepackage{adjustbox}
\usepackage{pgfplots}
\usepackage{gensymb}
\usepackage{booktabs}
\usepackage{siunitx}
\usepackage{etoolbox}
\usepackage{longtable}
\usepackage{bm}
\usepackage{overpic}
\usepackage{moresize}

\newcommand{\mAP}{\mathrm{mAP}}

\RequirePackage{xspace}
\makeatletter
\DeclareRobustCommand\onedot{\futurelet\@let@token\@onedot}
\def\@onedot{\ifx\@let@token.\else.\null\fi\xspace}
\def\eg{\emph{e.g}\onedot} 
\def\ie{\emph{i.e}\onedot} 
\def\cf{\emph{cf}\onedot}

\definecolor{customblue}{HTML}{313B72}
\definecolor{lightblue}{rgb}{0.9, 0.95, 1.0}
\definecolor{lightpruple}{HTML}{E1D5E7}

\newcommand{\smalltriangleright}{%
  \tikz[baseline=-0.7ex]{
    \fill[customblue, rounded corners=1.2pt] 
      (0,-0.8ex) -- (1.39ex,0) -- (0,0.8ex) -- cycle; 
  }%
  \hspace{0.2em}%
}

\usepackage[T1]{fontenc}
\usepackage{marvosym}

\begin{document}

\title{\LARGE \bf MIVIFI: Bridging Perspective and Fisheye Domains for Training Multi-View Fisheye Image Generation Models}

\author{%
    Matthias Neuwirth-Trapp$^{1,2,*,}$\thinspace\textsuperscript{\Letter},
    Begüm Altunbas$^{2,4,*}$,
    Jiayi Wang$^{2}$,
    Yan Xia$^{4}$,\\
    Maarten Bieshaar$^{2}$,
    Xinyu Huang$^{3}$,
    and Daniel Cremers$^{4}$%
    \thanks{$^{*}$Equal contribution. \textsuperscript{\Letter}\,mneuwirth@ethz.ch}%
    \thanks{$^{1}$ETH Zurich, Switzerland. \quad $^{2}$Bosch Research, Germany.}%
    \thanks{$^{3}$Bosch Research, USA. \quad $^{4}$Technical University of Munich, Germany.}%
}

\maketitle

\begin{abstract} Achieving 360° coverage is critical for the visual perception systems of autonomous vehicles. Fisheye cameras offer a cost-effective solution by enabling full surround coverage with as few as two sensors. However, existing multi-view fisheye datasets are limited, and synthesizing rare corner cases typically requires computationally expensive 3D simulations, hindering the training. While generative models have achieved significant success in standard perspective imagery, their application to wide-angle distortion remains unexplored. In this work, we formally introduce the novel problem of multi-view fisheye image generation conditioned on volumetric semantic representations and present two distinct methods. We first propose SyntheOcc-FE, which adapts the SyntheOcc architecture to fisheye data. While effective, this method is constrained by the scarcity of fisheye datasets, which limits its generalization. To overcome these limitations, we propose our second method, MIVIFI (\underline{m}ult\underline{i}-\underline{vi}ew \underline{fi}sheye), which leverages cross-domain learning with Equirectangular Projections. By bridging the gap between dataset domains using KITTI-360 fisheye images alongside nuScenes multi-view standard images, our approach enables high-fidelity manipulation of scene content. This framework enables the structural modification of semantic occupancy inputs to introduce or eliminate specific actors and facilitates the rendering of diverse meteorological conditions and illumination scenarios absent in the limited fisheye datasets. Quantitative and qualitative experiments demonstrate that our methods achieve robust photorealistic multi-view fisheye image generation and highlight the specific advantages of our cross-domain strategy for handling data scarcity. \end{abstract}
\section{Introduction}
\label{sec:introduction}
The robustness of perception systems in autonomous vehicles is fundamentally dependent on the quality and diversity of their training data. While surround-view systems are standard, fisheye cameras are increasingly adopted for their cost-effectiveness and expansive field of view~\cite{kumar2023surround}. This capability is critical for achieving comprehensive 360\degree\ coverage and eliminating blind spots during complex maneuvers like parking and navigating tight urban spaces. However, there is a pronounced scarcity of large-scale, richly annotated public datasets compared with their perspective-camera counterparts, such as nuScenes~\cite{nuscenes}.

\begin{figure}[t]
\centering
\includegraphics[width=1\linewidth]{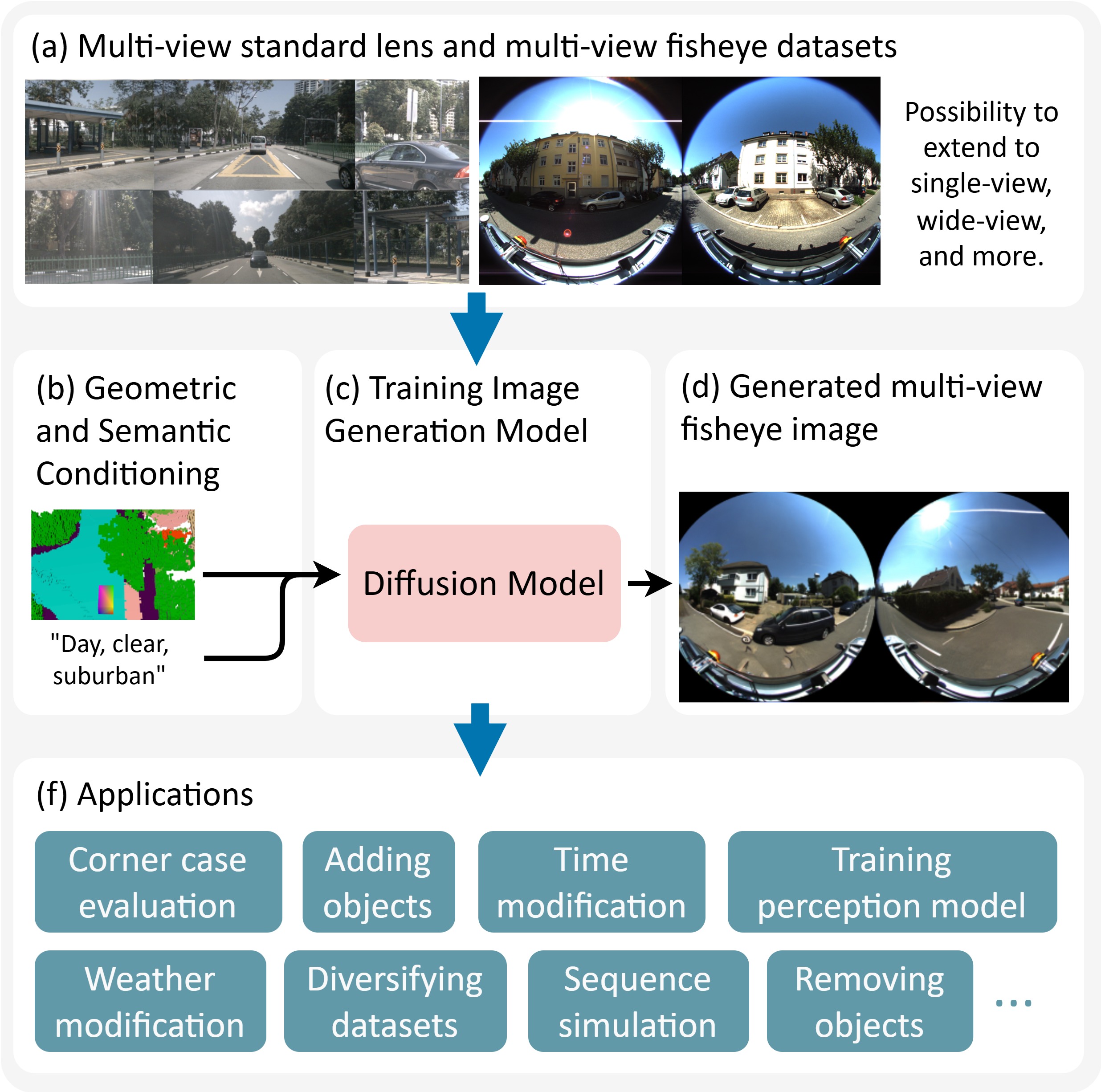}
\caption{\textbf{Controllable multi-view fisheye image generation.} We introduce the task of generating controllable multi-view fisheye images and present the first methods for it, \textbf{(1)~SyntheOcc-FE}, an adaptation of SyntheOcc~\cite{syntheocc} trained on fisheye datasets. To mitigate the limited availability of multi-view fisheye data and improve generalization, we further propose \textbf{(2) MIVIFI}, which builds on SyntheOcc-FE and uses equirectangular projections to enable training on both fisheye and standard-lens datasets (a). (b) Both methods are conditioned on semantic occupancy maps and text prompts and (c) train a diffusion model to synthesize multi-view fisheye images (d). The occupancy conditioning provides the image semantic layout, supporting various downstream applications (f).}

\label{fig:fig1}
\end{figure}

Recent advances in generative modeling, particularly diffusion-based frameworks such as Stable Diffusion~\cite{SD}, have shown promise for data augmentation and corner-case synthesis in autonomous driving~\cite{magicdrive}. Unlike text-to-image models, architectures conditioned on 3D geometric and semantic information, such as semantic occupancy grids, offer precise spatial control over the generated output~\cite{magicdrive,syntheocc,wovogen}. This paradigm enables rigorous scene editing: by manipulating the underlying volumetric semantic representation, one can systematically add or remove dynamic agents, alter infrastructure, or modify the scene layout before rendering the final image. Yet, applying these capabilities to multi-view fisheye imagery remains an unexplored and critical research gap. Direct application of existing methods is nontrivial due to severe lens distortion. Furthermore, a significant domain gap exists between visually diverse perspective-view datasets (\eg, nuScenes~\cite{nuscenes}) and more homogeneous fisheye datasets (\eg, KITTI-360~\cite{kitti360}), complicating knowledge transfer.

This paper addresses the scarcity of annotated multi-view fisheye datasets by proposing, to the best of our knowledge, the first two generation-based approaches for controllable multi-view fisheye images synthesis conditioned on 3D semantic occupancy grids. We first introduce \textbf{SyntheOcc-FE}, which adapts the SyntheOcc architecture to generate two $180\degree\times 180\degree$ fisheye images (\cf Figure~\ref{fig:fig1}). While this approach establishes a strong baseline for geometrically consistent generation, it is constrained by the limited diversity of existing fisheye data (\eg, KITTI-360), which consists of daytime, clear-weather scenes. This restricts the model's ability to synthesize critical long-tail scenarios, such as nighttime or rainy conditions, or to add new content.

To overcome these limitations and address the scarcity of diverse fisheye data, we propose a second novel method, \textbf{MIVIFI}, that leverages cross-domain training with fisheye and perspective multi-view images (\cf, Figure~\ref{fig:fig1}). By utilizing a shared Equirectangular Projection (ERP) space~\cite{depthanycamera}, MIVIFI integrates fisheye and standard-perspective images into a unified representation. This approach not only enables the transfer of visual attributes, such as weather and lighting effects, from the perspective domain to the fisheye domain but also enables advanced scene editing. We can structurally modify the input semantic occupancy, such as by inserting specific traffic participants or removing occlusions, and render these altered geometries under diverse atmospheric conditions absent from the original fisheye dataset. MIVIFI learns two wide-view ERP images that can be converted into two fisheye images, thereby achieving multi-view coverage.

\begin{figure*}[t]
    \centering
    \includegraphics[width=1\linewidth]{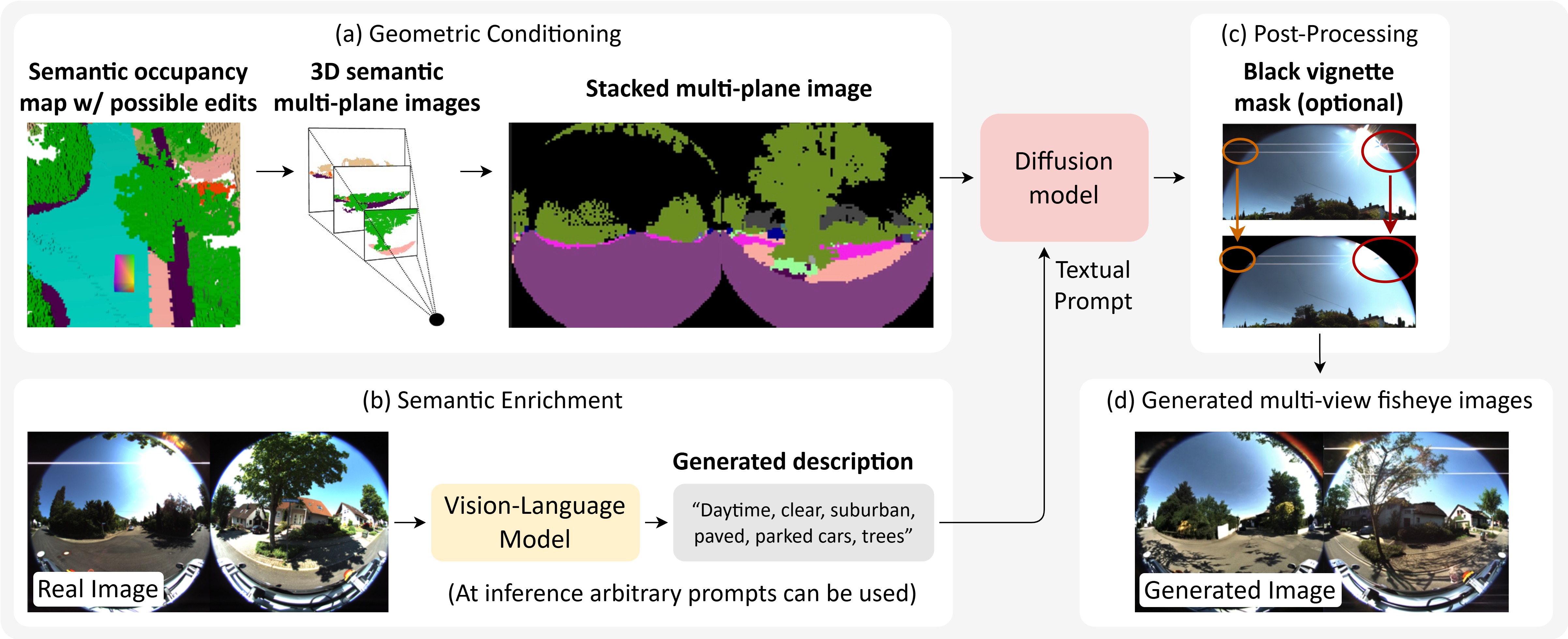}
\caption{\textbf{SyntheOcc-FE and MIVIFI overview.}
(a) \emph{Geometric conditioning:} a 3D semantic occupancy map is projected into a fisheye-aware multi-plane image (MPI) by discretizing depth and stacking the per-plane semantic layers (for each target view), yielding the structural control signal. 
(b) \emph{Semantic enrichment:} a vision-language model provides training captions, while arbitrary prompts are used at inference to steer appearance. The fisheye MPI and text prompt jointly condition a diffusion model to generate multi-view fisheye images (d). 
(c) \emph{Post-processing:} an optional binary vignette mask mimics mechanical lens vignetting and suppresses boundary artifacts. 
\textbf{MIVIFI} uses the same MPI+text conditioning but maps both fisheye and perspective data to a shared equirectangular projection (ERP) canvas, enabling joint training with large-scale perspective datasets via stitched ERP views and masked supervision in unobserved regions.}
    \label{fig:syntheocc-fe-arch}
\end{figure*}

Our experiments demonstrate that both of our methods generate high-quality images, with SyntheOcc-FE focusing on geometric fidelity and MIVIFI on generation diversity and semantic manipulation. For the first time, we can controllably synthesize corner cases, such as nighttime and rainy weather, for multi-view fisheye cameras by transferring knowledge from datasets in which these conditions were never captured with such sensors. This provides a scalable solution to the critical data scarcity problem for fisheye sensors, enabling the creation of diverse, edited datasets needed to train and validate safer, more reliable autonomous systems.

\noindent\textbf{In summary, our main contributions are:}
\begin{itemize}[leftmargin=*]
\item We formulate the novel task of generating controllable multi-view fisheye images conditioned on 3D semantic occupancy grids.
\item We introduce SyntheOcc-FE, a geometry-aware framework that adapts latent diffusion models to enforce strict fisheye lens characteristics and wide-angle consistency.
\item We propose MIVIFI, a unified training strategy that uses ERP to fuse fisheye and perspective datasets, enabling cross-domain transfer of diverse environmental semantics and facilitating precise structural editing of scene content.
\end{itemize}
\section{Related Work}
\label{sec:related_work}

\subsection{General Image Generation}
Early image synthesis relied on Generative Adversarial Networks (GANs)~\cite{GANs}, which often suffered from training instability and limited mode coverage. The advent of Denoising Diffusion Probabilistic Models (DDPMs)~\cite{DDPM} and Stable Diffusion (SD)~\cite{SD} revolutionized the field by enabling stable and diverse generation via iterative denoising in latent space. ControlNet~\cite{controlnet} further enhanced this by injecting spatial conditions into the generation process to guide structure.
While these foundational models enable high-fidelity synthesis, they inherently lack the geometric constraints required to model the nonlinear distortions of fisheye lenses without the explicit adaptation provided by our methods.

\subsection{Image Generation for Autonomous Driving}
In the autonomous driving domain, generative models have shifted toward multi-view consistency to simulate full vehicle surroundings. Methods like BEVGen~\cite{bevgen} and BEVControl~\cite{bevcontrol} utilize Bird's Eye View layouts to control scene synthesis, while SyntheOcc~\cite{syntheocc}, MagicDrive~\cite{magicdrive}, and WoVoGen~\cite{wovogen} employ 3D-aware conditioning to maintain geometric consistency across multiple cameras.
These approaches establish strong baselines for multi-view generation but are designed exclusively for perspective cameras. They fail to account for the severe non-linear distortion and wide field of view characteristic of fisheye sensors, which we explicitly model.

\subsection{Fisheye Image Generation}
Fisheye cameras are indispensable in modern autonomous vehicles for providing $360^\circ$ surround coverage and eliminating blind spots with minimal hardware. Despite their importance, generative research for this domain remains sparse. Curved Diffusion~\cite{curveddiffusion} introduced a lens-aware framework capable of simulating arbitrary optical geometries, but it does not focus on automotive applications. Similarly,~\cite{KG25} explored fisheye generation for edge-case detection but restricted their scope to static surveillance settings rather than dynamic driving environments. We address this gap by proposing the first framework specifically tailored for multi-view fisheye generation in autonomous driving which ensures both geometric accuracy and semantic consistency across overlapping wide-angle views.

\section{Methodology}
\label{sec:methodology}
In this section, we introduce SyntheOcc-FE and MIVIFI, the two methods proposed for multi-view fisheye generation.

\subsection{First Method: SyntheOcc-FE}
We propose SyntheOcc-FE, an adaptation of the SyntheOcc~\cite{syntheocc} framework tailored to the complex geometry of fisheye cameras. The pipeline consists of three stages: (1) geometric conditioning, (2) semantic enrichment, and (3) domain-aware fine-tuning.

First, we discretize the scene into Multi-Plane Images (MPI) with $N$ layers, which serve as the primary spatial conditioning signal. Unlike standard approaches, we modify the MPI generation to explicitly account for wide-angle distortion. Second, to capture visual diversity and scene semantics, we initialize our Stable Diffusion backbone (ControlNet and U-Net) with weights pre-trained on nuScenes and augment the training data with rich, vision-language-model-generated text descriptions. Finally, we fine-tune the model on our fisheye dataset to align the generative priors with the target sensor characteristics. The overall training setup is shown in Figure~\ref{fig:syntheocc-fe-arch}.

\vspace{0.5em}
\noindent \textbf{Geometric Conditioning via Fisheye MPI.}
To ensure the MPI conditioning is geometrically consistent with the sensor, we replace standard perspective projection with the Mei fisheye model~\cite{mei}. This allows us to accurately map semantic labels from the voxel grid to the distorted image plane (\cf Figure~\ref{fig:syntheocc-fe-arch} (a)). For a specific depth slice $d_i$, we first undistort a pixel's coordinates $(u, v)$ using the intrinsic matrix and distortion coefficients $(k_1, k_2, p_1, p_2)$ to obtain normalized coordinates $(x_n, y_n)$. These are mapped onto the unit sphere following the Mei projection:
\begin{equation}
    \mathbf{r} = 
    \begin{bmatrix}
        x_n \cdot c \\
        y_n \cdot c \\
        c - \xi
    \end{bmatrix},
\end{equation}
where $\xi$ represents the mirror parameter and the scalar $c$ is defined as:
\begin{equation}
    c = \frac{\xi \;\pm\; \sqrt{1 + (1-\xi^2)(x_n^2 + y_n^2)}}{1 + x_n^2 + y_n^2}.
\end{equation}
The resulting ray $\bm{r}$ is scaled by the target depth $d_i$ and transformed into world coordinates to sample the correct semantic label. This process ensures that the generated views preserve the specific non-linear distortion of the fisheye lens. Figure~\ref{fig:syntheocc-fe-arch} (a) shows an example of the final fisheye MPI, which is similar to a semantic segmentation but obtained from the 3D occupancy map.

\vspace{0.5em}
\noindent \textbf{Network Training and Post-Processing.}
With the geometric conditioning established, we leverage a Vision-Language Model (VLM) to annotate the fisheye dataset. These descriptions serve as text prompts that, combined with the MPI control signal, condition the diffusion model during fine-tuning. See Figure~\ref{fig:syntheocc-fe-arch} (b) for a VLM-generated sample caption. This could be further combined with prompt tuning approaches~\cite{jia2022visual, neuwirth2025incremental}.

As a final step to enforce sensor realism, we apply a binary mask to the generated output, setting all pixels within the mechanical vignetting area to black to match the physical constraints of the lens assembly (\cf Figure~\ref{fig:syntheocc-fe-arch} (c)). The general training pipeline is similar to SyntheOcc~\cite{syntheocc} and is schematically shown in Figure~\ref{fig:syntheocc-fe-arch}, in which the MPI of the occupancy map and the text prompt condition an image-generation model to obtain synthetic multi-view fisheye images.

\subsection{Second Method: MIVIFI}
To extend SyntheOcc-FE and incorporate the diversity of standard multi-view datasets (\eg, nuScenes), we introduce MIVIFI. While fisheye datasets offer structural realism, the available datasets often lack the semantic variety found in large-scale perspective collections. Building on SyntheOcc-FE, MIVIFI bridges this gap by adopting Equirectangular Projection (ERP) as a unified representation. By projecting both fisheye and perspective data onto a shared spherical manifold, we create a consistent domain for joint training. This allows the model to learn high-fidelity fisheye structures while benefiting from the rich object diversity of perspective data, conditioned on the same VLM-based text prompts used in SyntheOcc-FE, and optionally on the same circle mask to match the physical constraints of the lens assembly.

\vspace{0.5em}
\noindent \textbf{Unified ERP Representation.} We standardize all inputs into a common 180° × 180° Equirectangular Projection (ERP) layout. This projection maps the 3D spherical environment onto a flat 2D grid by treating longitude and latitude directly as horizontal and vertical image coordinates.

Wide-angle fisheye images map directly onto this expansive canvas, whereas standard perspective cameras with narrower fields of view yield sparse projections individually. To approximate wide-angle coverage, we stitch adjacent perspective cameras with overlapping sightlines into composite ERP patches. This strategy minimizes empty padding and ensures spatially comparable supervision across varying sensor types by masking any remaining unobserved regions. Furthermore, this masking approach allows the inclusion of single-view datasets for training, thereby significantly expanding the available data pool.

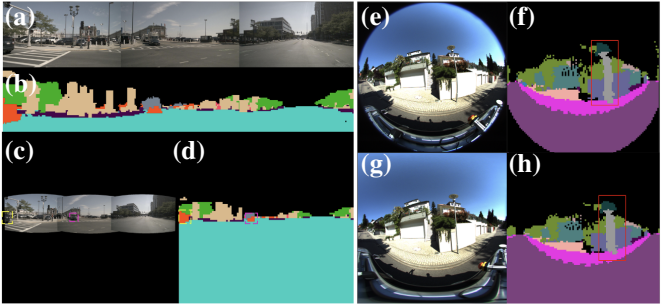
\begin{figure}[t]
    \centering

    \begin{overpic}[width=0.5\textwidth]{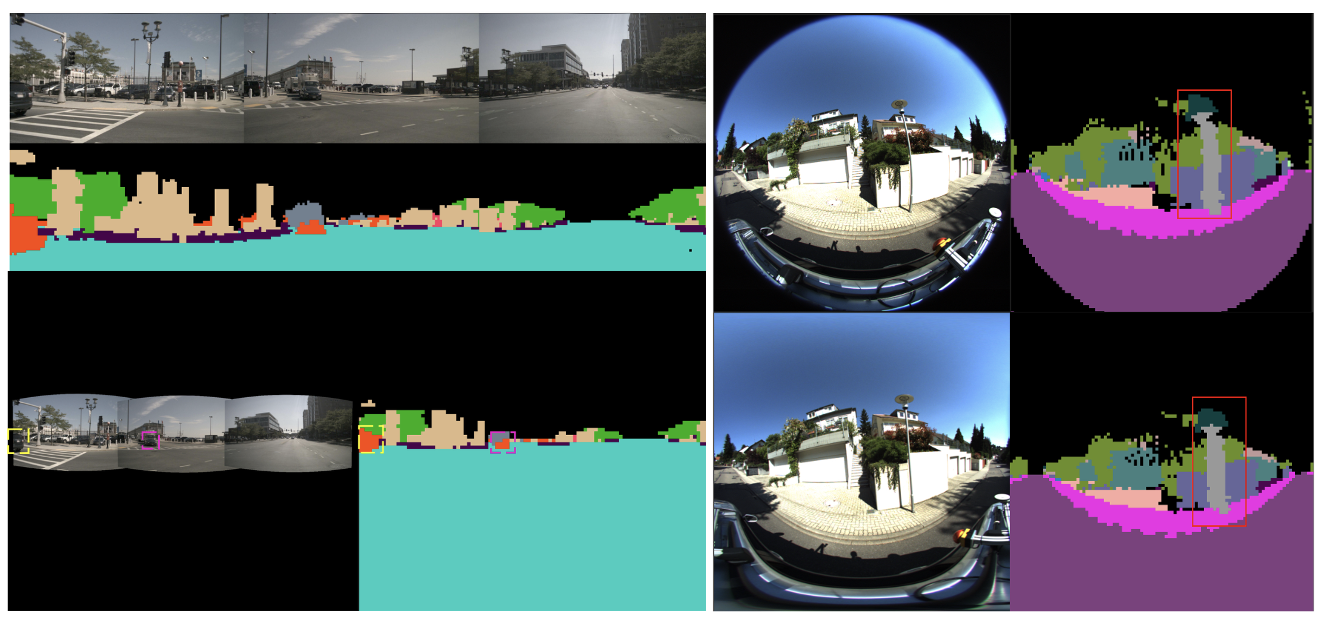}
        \put(1,43){\color{white}\textbf{(a)}}
        \put(1,33){\color{white}\textbf{(b)}}
        \put(1,23){\color{white}\textbf{(c)}}
        \put(26,23){\color{white}\textbf{(d)}}
        \put(54,43){\color{white}\textbf{(e)}}
        \put(76,43){\color{white}\textbf{(f)}}
        \put(54,21){\color{white}\textbf{(g)}}
        \put(76,21){\color{white}\textbf{(h)}}
    \end{overpic}
    \caption{\textbf{Unified ERP Data Preparation.} (a) Original perspective cameras and (b) their MPI. (c) Perspective cameras mapped to the ERP frame and (d) their corresponding MPI. (e) A fisheye image and (f) its MPI. (g) The fisheye image mapped to the ERP frame and (h) its corresponding MPI.}
    \label{fig:erp_version}
\end{figure}

For example, Figure~\ref{fig:erp_version} shows the direct mapping of our wide-angle fisheye images onto the full ERP canvas, while stitching three adjacent perspective cameras from nuScenes to form a comparable wide-field representation.

\vspace{0.5em}
\noindent \textbf{Masked Supervision and Out-of-FoV Completion.}
Despite stitching, perspective ERP projections inevitably contain unobserved regions (padding). To prevent the model from overfitting to these artifacts, we apply a masked loss that restricts supervision to valid pixel regions. Conversely, fisheye images provide valid signals across the entire canvas. 

The bottom row in Figure~\ref{fig:erp_version} shows on the left the stitched perspective images with large black background sections, for which the training signal will be removed. Crucially, while the RGB input for perspective data is masked, we project the corresponding occupancy MPIs across the full $180^\circ \times 180^\circ$ domain. This creates a training setup in which the model is supervised by complete structural information (MPI), even when visual information (RGB) is missing. This implicitly encourages the model to perform out-of-FoV completion, \ie, hallucinating consistent RGB content in empty regions by leveraging cross-domain knowledge transfer from fully realized fisheye examples.

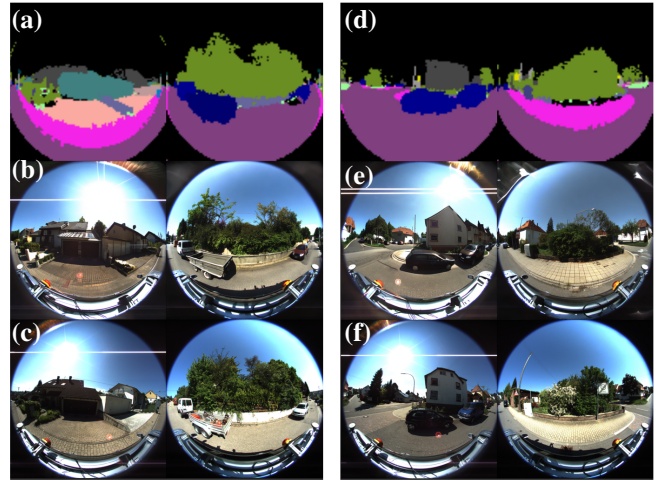
\begin{figure}[t]
    \centering
            \begin{overpic}[width=1\linewidth]{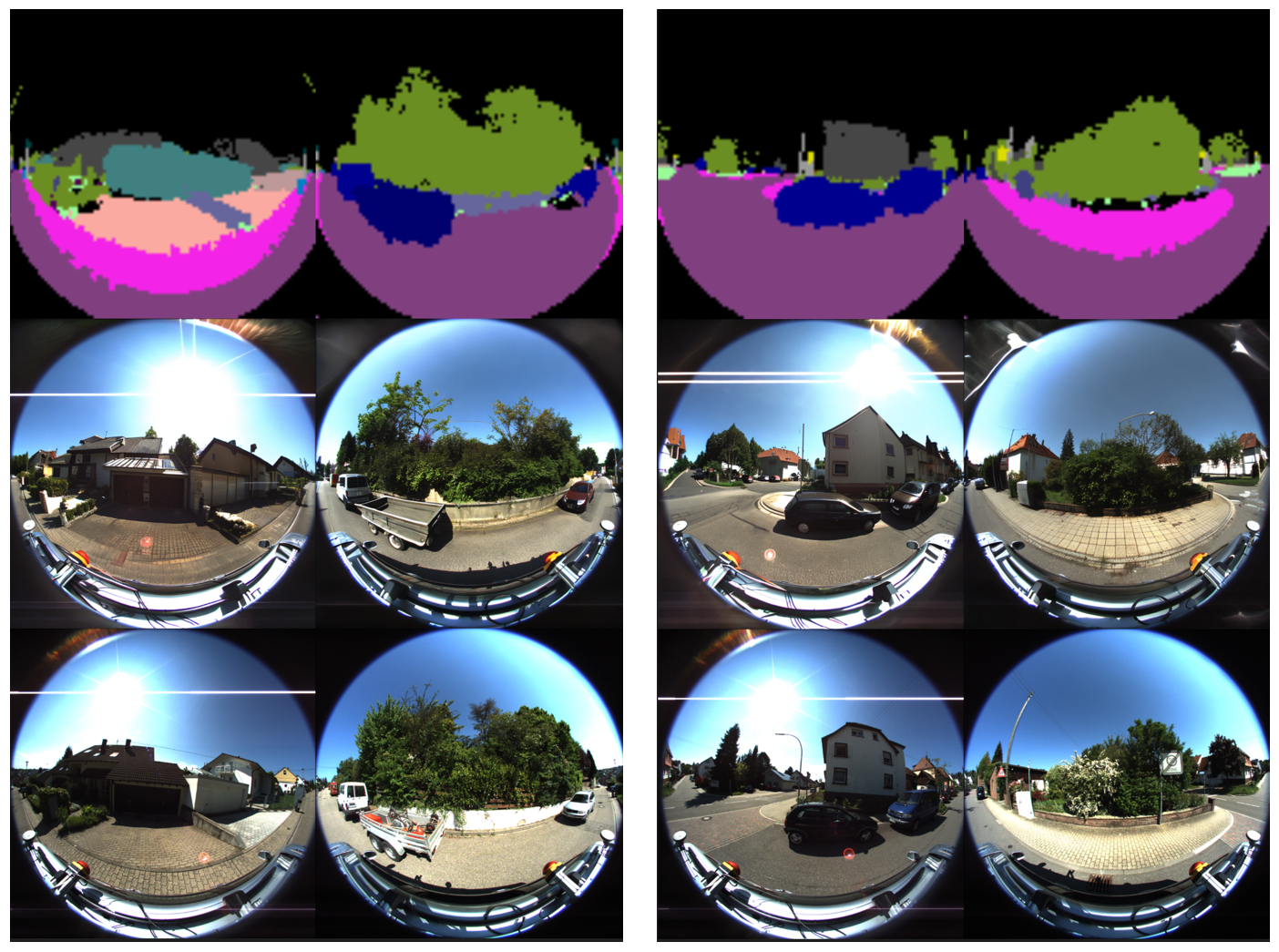}
                \put(1,70){\color{white}\textbf{(a)}}
                \put(1,47){\color{white}\textbf{(b)}}
                \put(1,22){\color{white}\textbf{(c)}}
                \put(52,70){\color{white}\textbf{(d)}}
                \put(52,46){\color{white}\textbf{(e)}}
                \put(52,22){\color{white}\textbf{(f)}}
            \end{overpic}
\caption{\textbf{Qualitative Results of SyntheOcc-FE.} (a, d) Input MPI, (b, e) generated fisheye images, and (c, f) the corresponding real multi-view fisheye images for reference.}
    \label{fig:qualitative_comparison_SyntheOcc-FE}
\end{figure}

\begin{figure}[t]
    \centering
    \begin{subfigure}[c]{0.48\linewidth}
        \centering
        \resizebox{\linewidth}{!}{
        \adjustbox{clip, trim=0pt 0pt {0.5\width} 0pt}{
            \begin{overpic}{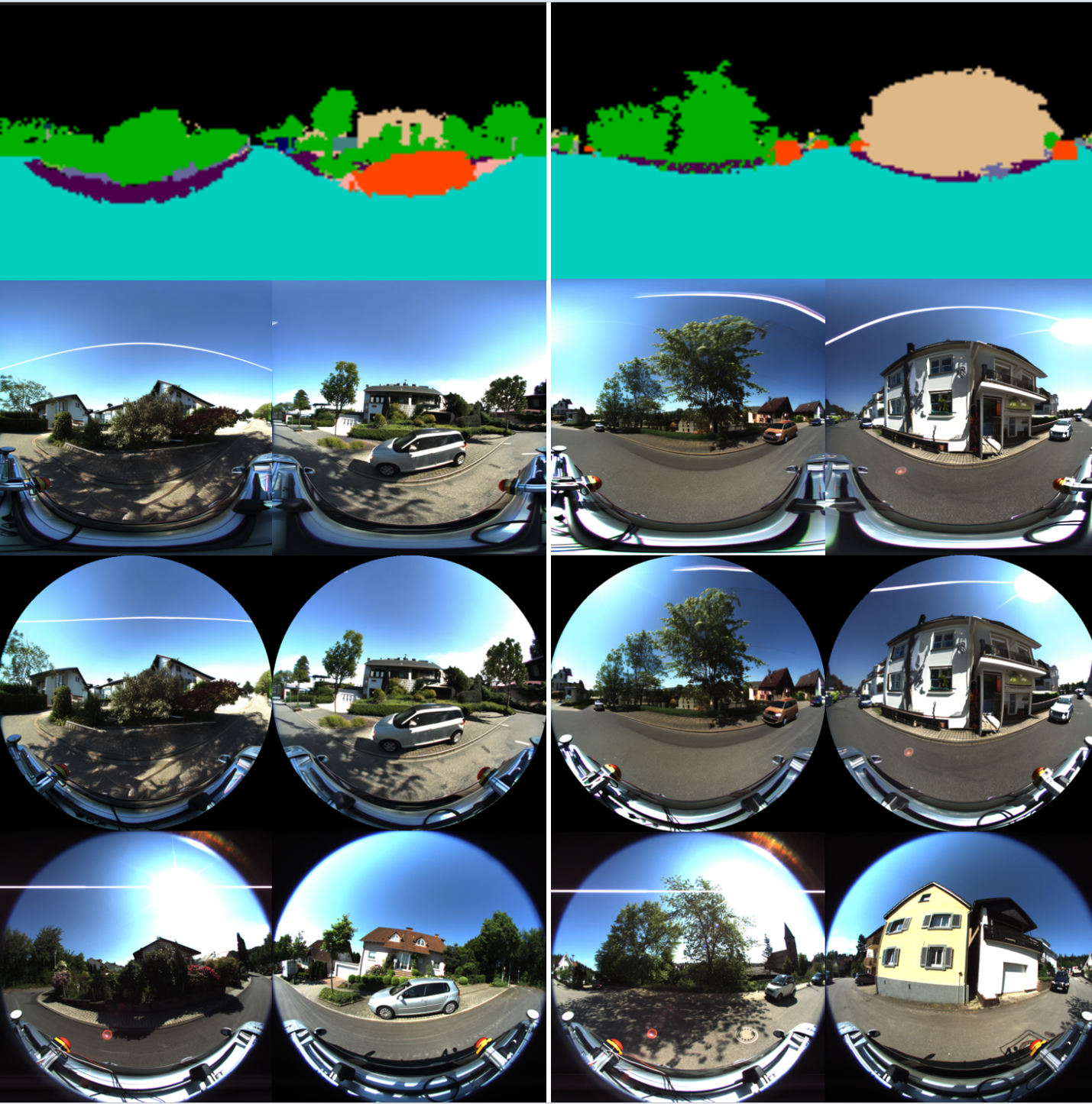}
                \put(1,96){\color{white}\scalebox{1.3}{\HUGE\textbf{(a)}}}
                \put(1,71){\color{white}\scalebox{1.3}{\HUGE\textbf{(b)}}}
                \put(1,46){\color{white}\scalebox{1.3}{\HUGE\textbf{(c)}}}
                \put(1,21){\color{white}\scalebox{1.3}{\HUGE\textbf{(d)}}}
            \end{overpic}
        }
        }
    \end{subfigure}
    \hfill
    \begin{subfigure}[c]{0.48\linewidth}
        \centering
        \resizebox{\linewidth}{!}{

                    \adjustbox{clip, trim=0pt 0pt {0.5\width} 0pt}{
            \begin{overpic}{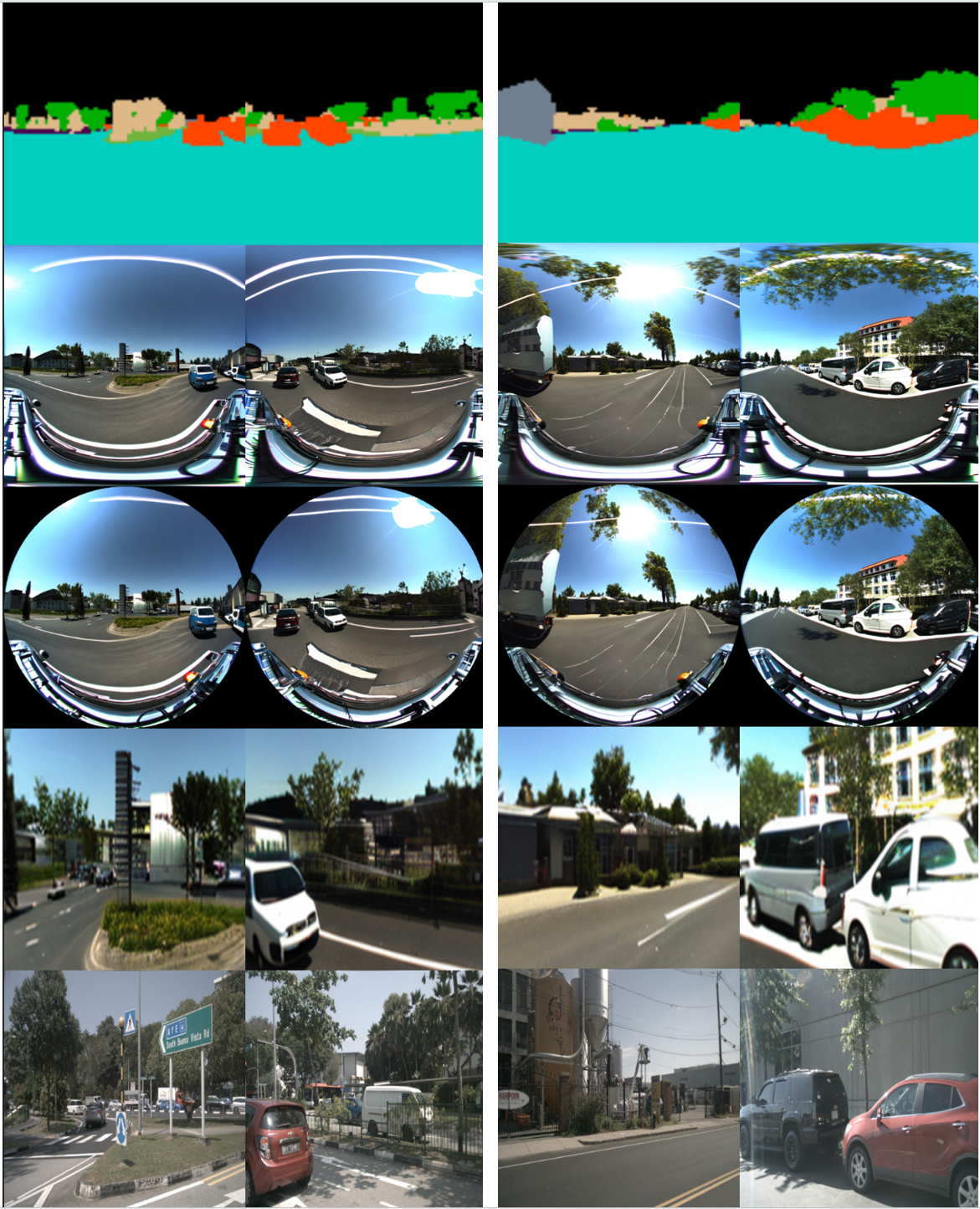}
                \put(1,96){\color{white}\scalebox{1.3}{\HUGE\textbf{(e)}}}
                \put(1,76){\color{white}\scalebox{1.3}{\HUGE\textbf{(f)}}}
                \put(1,56){\color{white}\scalebox{1.3}{\HUGE\textbf{(g)}}}
                \put(1,36){\color{white}\scalebox{1.3}{\HUGE\textbf{(h)}}}
                \put(1,16){\color{white}\scalebox{1.3}{\HUGE\textbf{(i)}}}
            \end{overpic}
        }
        }
    \end{subfigure}

    \vspace{8pt}

\caption{\textbf{Qualitative Results of MIVIFI.} Comparison of generated outputs for KITTI-360 (a)--(d) and nuScenes (e)--(i). Both datasets display (a, e) the ERP MPI, (b, f) the generated ERP, and (c, g) the ERP converted to fisheye. For comparison, KITTI-360 includes (d) the real fisheye image, while nuScenes includes (h) the converted perspective and (i) the real reference image. These results demonstrate visual consistency, high quality, and faithful input adherence, as well as highlighting the ability to convert perspective images into multi-view fisheye images covering a $360\degree \times 180\degree$ FoV.}
    \label{fig:qualitative_comparison_mivifi}
\end{figure}
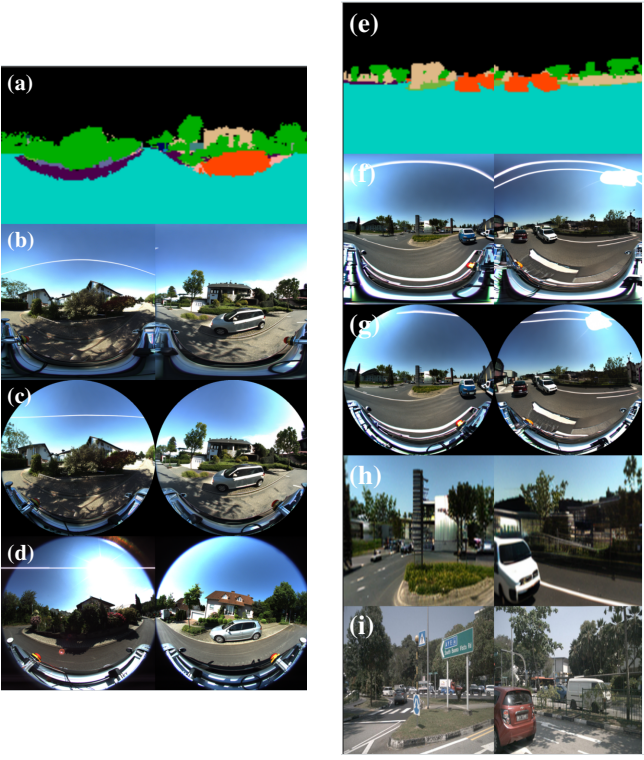

\section{Experiments}
\label{sec:experiments}

\subsection{Datasets}
\label{subsec:datasets}
Our experiments leverage KITTI-360~\cite{kitti360} for fisheye training and evaluation, and nuScenes~\cite{nuscenes} to broaden scene diversity via ERP-based joint training in MIVIFI. \textbf{KITTI-360} extends the original KITTI~\cite{kitti} benchmark with suburban driving sequences recorded in Germany and provides left- and right-facing fisheye cameras, each offering a $180\degree \times 180\degree$ field-of-view. Dense accumulated LiDAR scans are converted into semantic occupancy voxel grids at $0.2\,\text{m}$ resolution, which serve as our primary conditioning input. \textbf{nuScenes} comprises urban scenes from Boston and Singapore with six perspective cameras at approximately $70\degree \times 40\degree$ field-of-view; for this dataset, we adopt the dense occupancy voxel grids released by SurroundOcc~\cite{surroundocc} at $0.5\,\text{m}$ resolution.

\subsection{Implementation Details}
\label{subsec:implementation_details}
We build our model upon Stable Diffusion v2.1~\cite{SD} and represent occupancy as MPI of size $100 \times 100 \times 256$. Training utilizes a base batch size of 4 with gradient accumulation over 8 steps for an effective batch size of 32. To ensure balanced contributions to MIVIFI, we employ a batch sampler that selects equal numbers of samples from KITTI-360 and nuScenes. We set the learning rate to $1 \times 10^{-6}$. For inference, we use the UniPC scheduler~\cite{unipc} with a classifier-free guidance scale of 7.0 and 20 denoising steps.

\subsection{Metrics}
\label{subsec:metrics}
We assess quality, semantics, and diversity using four standard metrics within the valid fisheye region:

\begin{itemize}[leftmargin=*,itemsep=0.25em,topsep=0.25em]
  \item \textbf{FID:} The Fr\'echet Inception Distance~\cite{fid} measures the distance between real and generated image distributions; lower values imply higher fidelity.
  \item \textbf{SSIM:} The Structural Similarity Index~\cite{ssim} quantifies perceptual similarity in terms of luminance, contrast, and structure.
  \item \textbf{LPIPS:} We use Learned Perceptual Image Patch Similarity~\cite{lpips} to evaluate diversity. We compute pairwise distances between multiple images generated from the same input using different random seeds, with higher scores indicating greater variation.
  \item \textbf{mIoU:} We use mean Intersection over Union (mIoU) to assess semantic accuracy by comparing the semantic segmentation of generated and corresponding real images. For reliability, we filter out the ego-vehicle and mechanical lens vignette, and we exclude areas without an occupancy map due to weak LiDAR coverage, as these regions provide insufficient conditioning signal.
\end{itemize}

\subsection{Qualitative Results}\label{subsubsec:qualitative}

\paragraph{General Generation Quality}
Figure~\ref{fig:qualitative_comparison_SyntheOcc-FE} and~\ref{fig:qualitative_comparison_mivifi} show the qualitative results of SyntheOcc-FE and MIVIFI, respectively. Both approaches generate high-quality images that respect the given conditioning. Because only the MPI with semantic and spatial information and the textual prompt with an abstract description were provided, it is to be expected that the generated images differ slightly from the ground truth. This is further desirable, as it allows the generation of diverse datasets with the same semantic layout. We excluded the lens mask for the results here to show that the model, in general, learns to set these areas to black. We further want to highlight the out-of-FoV completion in~\ref{fig:qualitative_comparison_mivifi} (g). The generated fisheye images cover a larger area, whereas the nuScenes dataset contains only images of the central region. Compare to Figure~\ref{fig:erp_version} to see how many pixels are generated and do not have a real counterpart.

\begin{figure}[tb]
    \centering
    \begin{subfigure}[c]{0.48\linewidth}
        \centering
        \resizebox{\linewidth}{!}{
        \adjustbox{clip, trim=13pt 27pt {0.796\width} 2pt}{
            \begin{overpic}{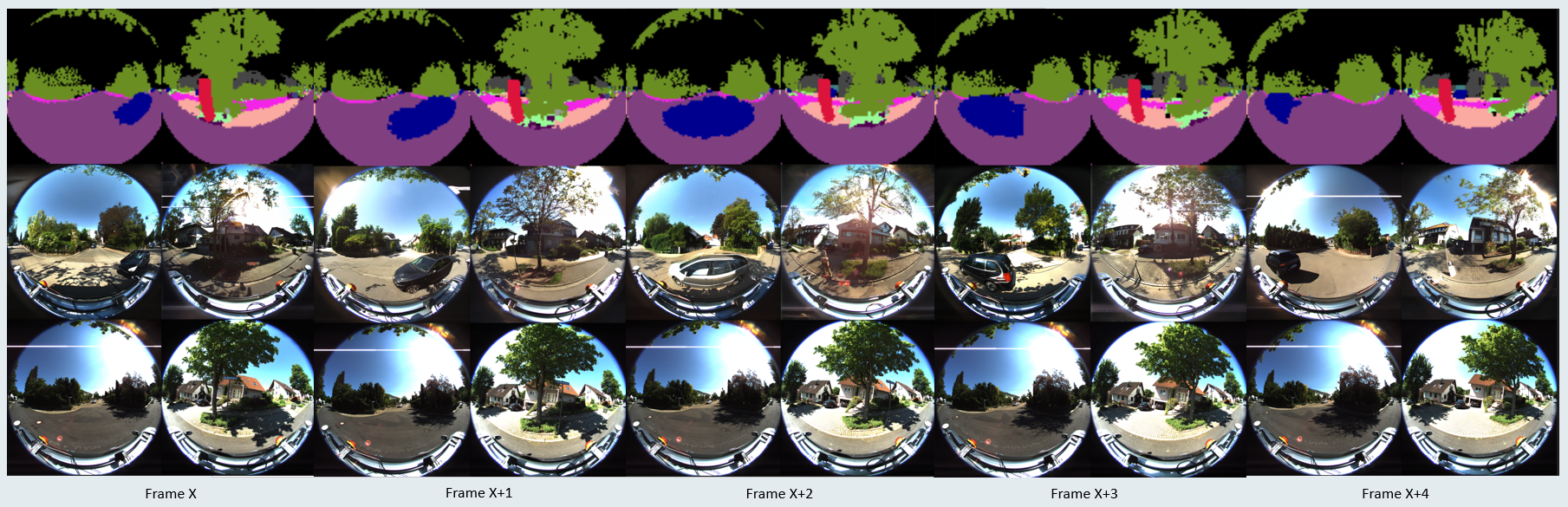}
                \put(0.7,30.8){\color{white}\scalebox{1}{\Large\textbf{(a)}}}
                \put(0.7,20.8){\color{white}\scalebox{1}{\Large\textbf{(b)}}}
                \put(0.7,10.8){\color{white}\scalebox{1}{\Large\textbf{(c)}}}
                \put(10,30.8){\color{white}\scalebox{1}{\Large\textbf{(d)}}}
                \put(10,20.8){\color{white}\scalebox{1}{\Large\textbf{(e)}}}
                \put(10,10.8){\color{white}\scalebox{1}{\Large\textbf{(f)}}}
            \end{overpic}
        }
        }
    \end{subfigure}
    \hfill
    \begin{subfigure}[c]{0.48\linewidth}
        \centering
        \begin{overpic}[width=1\linewidth]{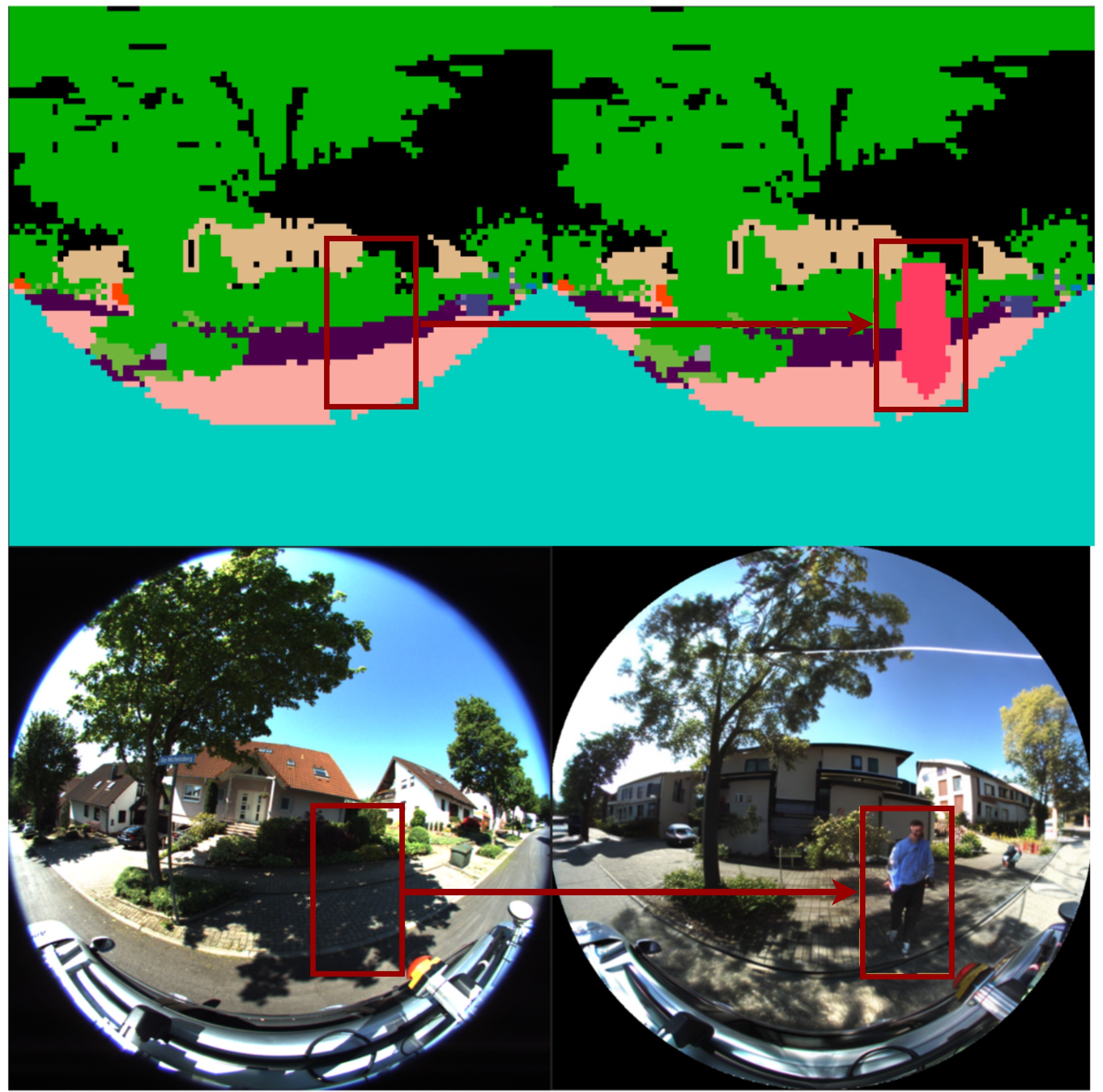}
            \put(1,92.5){\color{white}\scalebox{1}{\footnotesize\textbf{(g)}}}
            \put(1,44){\color{white}\footnotesize\textbf{(h)}}
            \put(51,92.5){\color{white}\footnotesize\textbf{(i)}}
            \put(51,44){\color{white}\footnotesize\textbf{(j)}}
        \end{overpic}
    \end{subfigure}

    \vspace{8pt}

    \caption{\textbf{Comparison of Object Insertion with SyntheOcc-FE and MIVIFI.} \textbf{Left:} Results using SyntheOcc-FE. (a, d) Modified MPIs for the left and right views, showing an inserted car (blue) and person (red). (b, e) The respective generated images, demonstrating successful car insertion (b) but failed person insertion (e). (c, f) The original multi-view reference images before modification. \textbf{Right:} Results using MIVIFI. (g) Original MPI and (h) the corresponding real image. (i) Modified MPI with an inserted person (red) and (j) the generated output. The expanded training dataset allows MIVIFI to successfully synthesize pedestrians.}
    \label{fig:combined_object_insertion}
\end{figure}
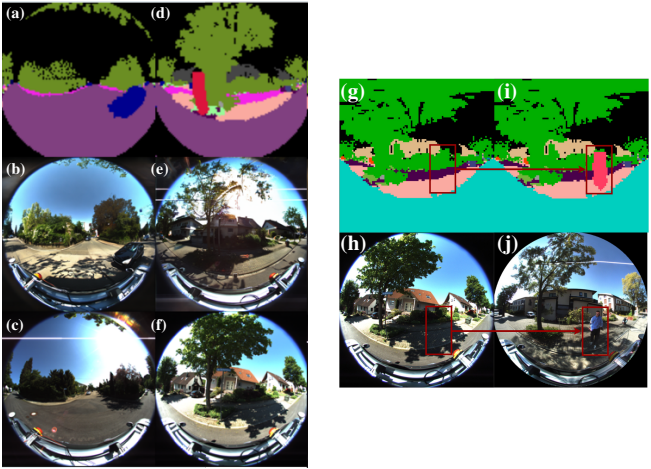

\paragraph{Adding New Objects}
We evaluate the insertion of new objects by modifying the underlying occupancy map. Figure~\ref{fig:combined_object_insertion} (a--f) illustrates the addition of a car and a person using SyntheOcc-FE. The model successfully generates a car that remains consistent with the MPI. However, the person (highlighted in red) is absent from the generated output. This limitation stems from the KITTI-360 training data, which lacks pedestrian instances.

To address this issue, we utilize MIVIFI, which incorporates the nuScenes dataset during training. This extension enables the successful generation of pedestrians as demonstrated in Figure~\ref{fig:combined_object_insertion} (g--j). These experiments confirm that scenes can be effectively manipulated by adding or removing objects within the occupancy map.

\begin{figure}[tb]
    \centering
    \begin{subfigure}[b]{0.48\linewidth}
        \centering
        \resizebox{\linewidth}{!}{
            \adjustbox{clip, trim=0pt 0pt 0pt {0.33\height}}{
                \begin{overpic}{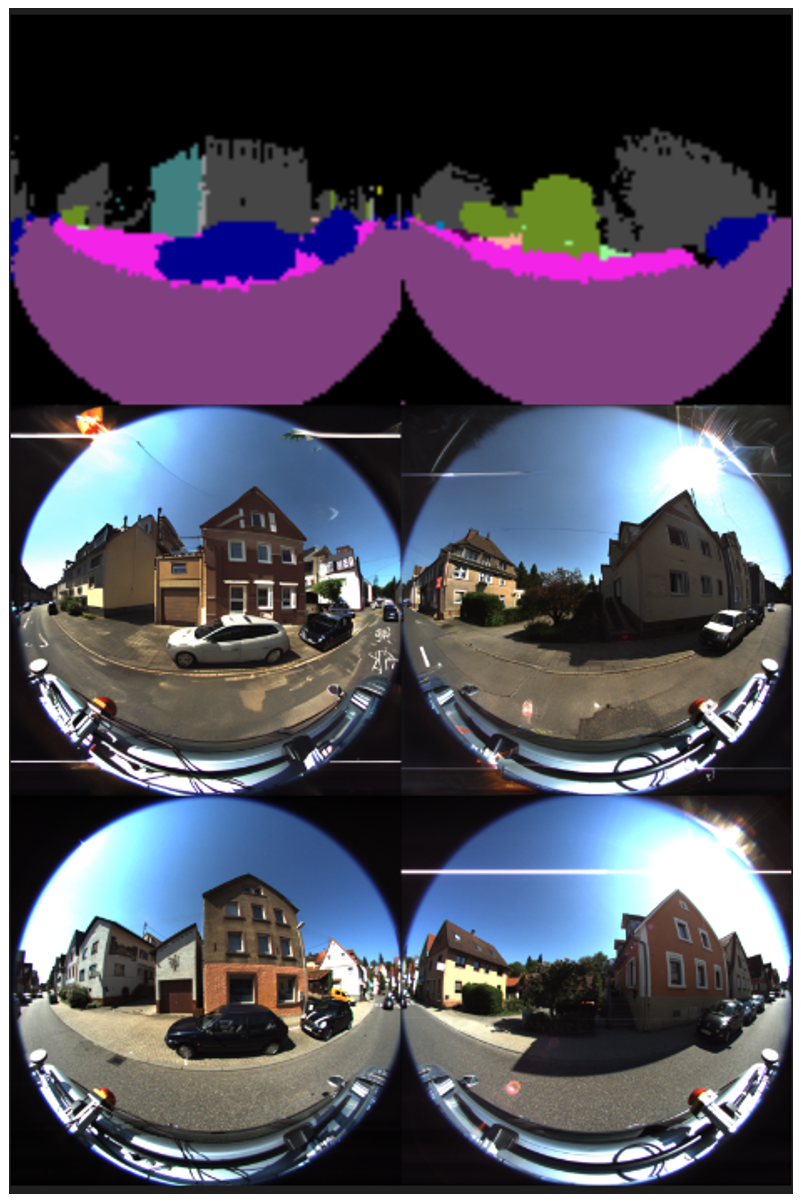}
                    \put(1.5,63){\color{white}\scalebox{1.3}{\HUGE\textbf{(a)}}}
                    \put(32,63){\color{white}\scalebox{1.3}{\HUGE\textbf{(b)}}}
                    \put(1.5,30){\color{white}\scalebox{1.3}{\HUGE\textbf{(c)}}}
                    \put(32,30){\color{white}\scalebox{1.3}{\HUGE\textbf{(d)}}}
                \end{overpic}
            }
        }
    \end{subfigure}
    \hfill
    \begin{subfigure}[b]{0.49\linewidth}
        \centering
        \resizebox{\linewidth}{!}{
            \adjustbox{clip, trim=0pt 0pt {0.5\width} {0.5\height}}{
                \begin{overpic}{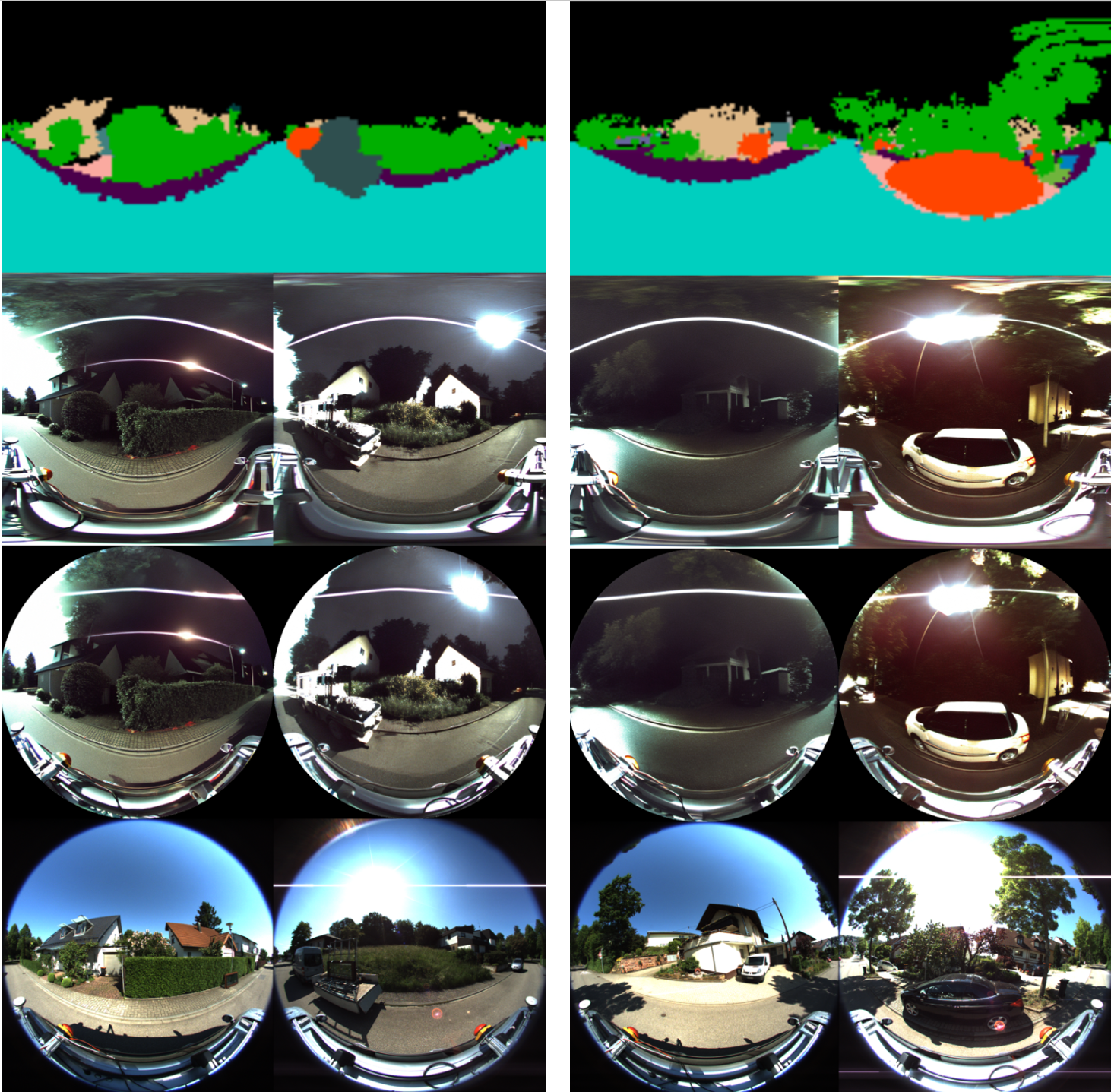}
                    \put(0.8,46){\color{white}\scalebox{1.2}{\HUGE\textbf{(e)}}}
                    \put(23,46){\color{white}\scalebox{1.2}{\HUGE\textbf{(f)}}}
                    \put(0.8,22){\color{white}\scalebox{1.2}{\HUGE\textbf{(g)}}}
                    \put(23,22){\color{white}\scalebox{1.2}{\HUGE\textbf{(h)}}}
                \end{overpic}
            }
        }
    \end{subfigure}
    \vspace{8pt}
    \caption{\textbf{Nighttime Synthesis Comparison.} Transformation of daytime multi-view fisheye imagery via text-prompt adaptation, comparing SyntheOcc-FE (a)--(d) and MIVIFI (e)--(h). (a, b) and (e, f) show the generated multi-view fisheye images conditioned by including ``nighttime'' in the prompt, while (c, d) and (g, h) are the real daytime images for comparison.}
    \label{fig:nighttime-comparison}
\end{figure}
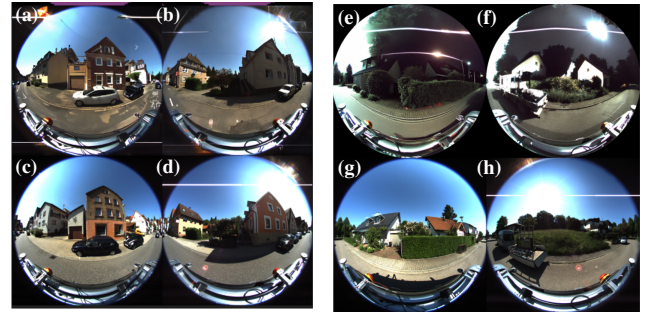

\paragraph{Shifting the Time-of-Day}
We control visual properties through text prompts by combining the time-agnostic MPI with textual descriptors such as ``nighttime.'' Figure~\ref{fig:nighttime-comparison} compares the performance of SyntheOcc-FE and MIVIFI when transforming a daytime scene. SyntheOcc-FE fails to synthesize night imagery due to limited training data in this modality. In contrast, MIVIFI leverages a broader training distribution from nuScenes to generate realistic nocturnal environments.



\begin{table*}[t]
\centering
\small
\renewcommand{\arraystretch}{1.1}
\caption{\textbf{Image quality results for SyntheOcc-FE and MIVIFI.} SyntheOcc-FE demonstrates superior generation quality, while MIVIFI achieves greater diversity. Best results are in bold.}
\label{tab:results}
\begin{tabular}{l c c c ccc c}
\toprule
& & & & \multicolumn{3}{c}{\textbf{mIoU Classes  $\uparrow$}} & \\
\cmidrule(lr){5-7}
\textbf{Method} & \textbf{FID $\downarrow$} & \textbf{SSIM $\uparrow$} & \textbf{mIoU $\uparrow$} & \textbf{Road  $\uparrow$} & \textbf{Car  $\uparrow$} & \textbf{Background  $\uparrow$} & \textbf{LPIPS $\uparrow$} \\
\midrule
\textbf{SyntheOcc-FE} (ours)   & \textbf{10.77} & \textbf{0.49} & \textbf{0.34} & \textbf{0.65} & \textbf{0.49} & \textbf{0.74} & 0.30 \\
\textbf{MIVIFI} (ours)       & 17.05 & 0.41 & 0.27 & 0.56 & 0.35 & 0.64 & \textbf{0.36} \\
\bottomrule
\end{tabular}
\end{table*}

\begin{figure}[tb]
    \centering
    \begin{subfigure}[b]{0.48\linewidth}
        \centering
        \resizebox{\linewidth}{!}{%
            \adjustbox{clip, trim=0pt 0pt {0.5\width} {0.33\height}}{%
                \begin{overpic}{figures/ft-llava.png}
                    \put(1.5,46.7){\color{white}\scalebox{1.1}{\HUGE\textbf{(a)}}}
                    \put(25,46.7){\color{white}\scalebox{1.1}{\HUGE\textbf{(b)}}}
                    \put(1.5,22.5){\color{white}\scalebox{1.1}{\HUGE\textbf{(c)}}}
                    \put(25,22.5){\color{white}\scalebox{1.1}{\HUGE\textbf{(d)}}}
                \end{overpic}%
            }%
        }
    \end{subfigure}
    \hfill
    \begin{subfigure}[b]{0.48\linewidth}
        \centering
        \resizebox{\linewidth}{!}{%
            \adjustbox{clip, trim=0pt 0pt {0.5\width} {0.5\height}}{%
                \begin{overpic}{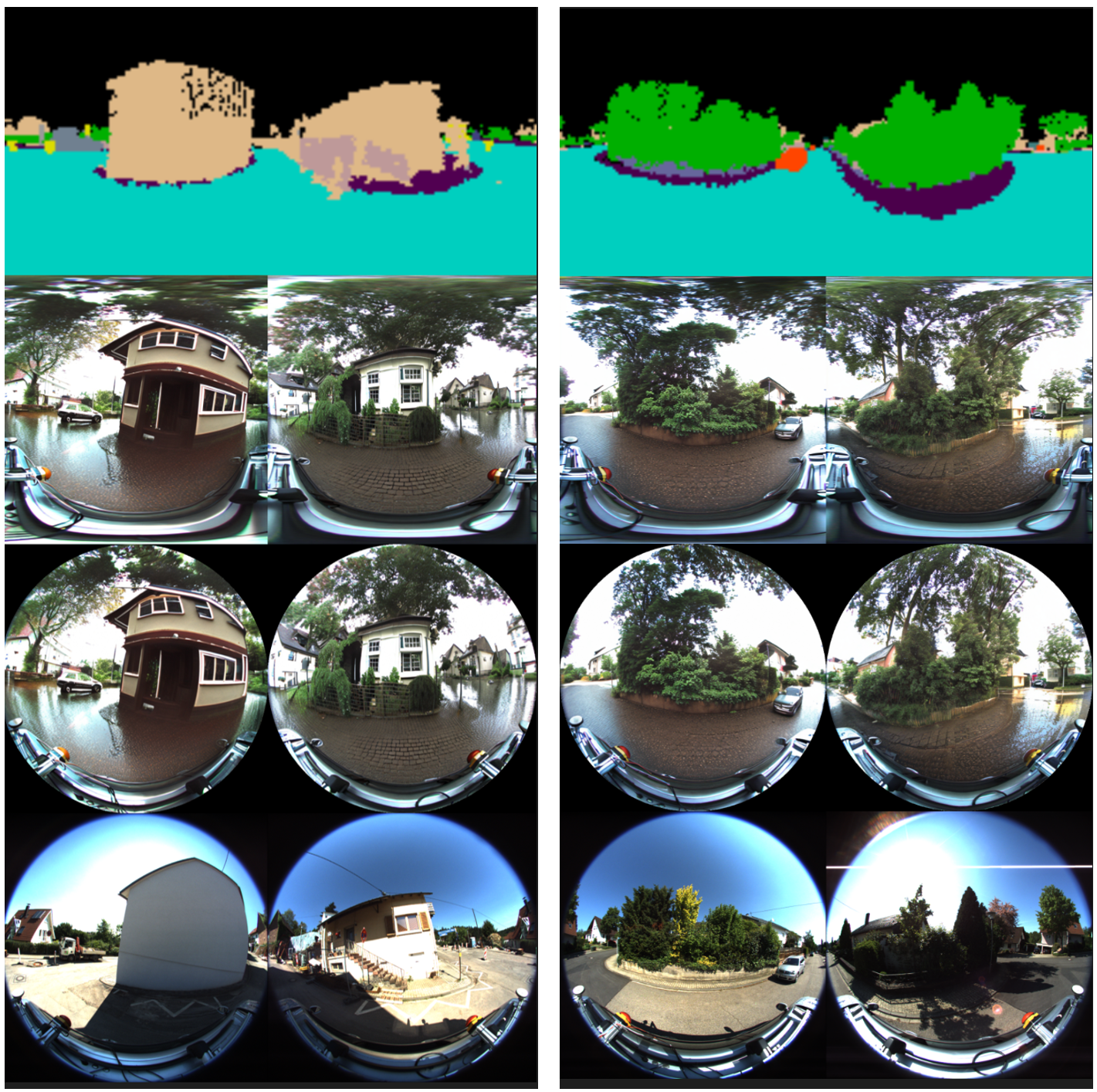}
                    \put(0.8,47){\color{white}\scalebox{1.1}{\HUGE\textbf{(e)}}}
                    \put(23,47){\color{white}\scalebox{1.1}{\HUGE\textbf{(f)}}}
                    \put(0.8,23){\color{white}\scalebox{1.1}{\HUGE\textbf{(g)}}}
                    \put(23,23){\color{white}\scalebox{1.1}{\HUGE\textbf{(h)}}}
                \end{overpic}%
            }%
        }
    \end{subfigure}
    \vspace{8pt}
    \caption{\textbf{Rain Synthesis Comparison.} Transformation of sunny multi-view fisheye imagery via text-prompt adaptation to rain, comparing SyntheOcc-FE (a)--(d) and MIVIFI (e)--(h). (a, b) and (e, f) show the generated multi-view fisheye images conditioned by including ``rain'' in the prompt, while (c, d) and (g, h) are the real sunny images for comparison. SyntheOcc-FE is unable to generate rain, while MIVIFI successfully renders the weather modification realistic.}
    \label{fig:weather-comparison}
\end{figure}

\paragraph{Modifiying the Weather}
We further manipulate environmental attributes via text prompts by integrating the time-agnostic MPI with specific weather descriptors, such as ``heavy rain.'' Figure~\ref{fig:weather-comparison} illustrates the contrast between SyntheOcc-FE and MIVIFI when adapting a clear-weather scene. While SyntheOcc-FE struggles to generate plausible precipitation effects due to data sparsity in this domain, MIVIFI leverages a more comprehensive training distribution provided by nuScenes.

\subsection{Quantitative Results}
We present quantitative results using the metrics described in Section~\ref{subsec:metrics}. Table~\ref{tab:results} compares SyntheOcc-FE and MIVIFI. As SyntheOcc-FE is specialized for fisheye images, it achieves higher scores across all quality-based metrics. In contrast, MIVIFI emphasizes generalization and diversity, as evidenced by lower quality but higher LPIPS scores.

Table~\ref{tab:downstream_task} compares methods on a downstream 2D object detection task. Following the COCO protocol, we report mean average precision over multiple intersection-over-union thresholds, defined as $\mathrm{mAP}=\mathrm{mAP}@\text{[.5:.05:.95]}$~\cite{fleet_microsoft_2014}. We trained YOLOv11~\cite{yolo11_ultralytics} on 4,000 images from the real KITTI-360 dataset or generated samples. We test it on the KITTI-360 test dataset. While real data outperforms synthetic, synthetic results remain competitive. SyntheOcc-FE generally yields better results than MIVIFI. However, because KITTI-360 is scarce in pedestrians, MIVIFI demonstrates the benefit of diverse training data for this specific class. Overall, both generative approaches yield viable results but fall short of the real training data.

\subsection{Ablations}

We evaluate several architectural modifications to analyze their impact on performance (\cf Table~\ref{tab:SyntheOcc-FE_ablations}).

\begin{itemize}
\item Replacing VLM-generated prompts with generic prompts significantly degrades image quality, although Background mIoU remains high due to the broad nature of that category.
\item Removing the nuScenes-pretrained weights in addition to using generic prompts further reduces performance.
\item Eliminating the lens mask increases the FID to 18.68.
\item Conditioning image generation on a pseudo-depth map from Depth Any Camera~\cite{depthanycamera}, cropped at the top and bottom to exclude unreliable regions, slightly increases the FID despite providing additional geometric guidance alongside the MPI.
\item Incorporating a cosine-similarity loss to improve overlap consistency also increases the FID, indicating that the optimization prioritizes loss minimization over overall image quality.
\end{itemize}

\begin{table*}[t]
    \centering
    \small
    \renewcommand{\arraystretch}{0.94}
    \setlength{\tabcolsep}{3pt}
    \caption{\textbf{Ablation Study on SyntheOcc-FE.} Removing VLM prompts and lens masks degrades overall performance. Conversely, introducing depth conditioning and multi-view cosine similarity yields no significant quantitative gains. SSIM and mIoU are overall more stable than FID. Best in bold.}
    \label{tab:SyntheOcc-FE_ablations}
    \begin{tabular}{l c c c ccc c}
        \toprule
        & & & & \multicolumn{3}{c}{\textbf{mIoU Classes} $\uparrow$} & \\
        \cmidrule(lr){5-7}
        \textbf{Experiment} & \textbf{FID} $\downarrow$ & \textbf{SSIM} $\uparrow$ & \textbf{mIoU} $\uparrow$ & \textbf{Road} & \textbf{Car} & \textbf{Backgr.}  \\
        \midrule
        SyntheOcc-FE (ours) & \textbf{10.77} & \textbf{0.49} & \textbf{0.34 }& \textbf{0.65} & 0.49 & \textbf{0.74}  \\

        SyntheOcc-FE (ours) w/o VLM prompts & 21.25 & 0.48 & 0.33 & 0.55 & 0.45 & \textbf{0.74} \\
        SyntheOcc-FE (ours) w/o VLM prompts \& w/o nuScenes-pretrained weights & 23.32 & 0.40 & 0.28 & 0.47 & 0.36 & 0.72\\
        SyntheOcc-FE (ours) w/ 2D depth condition & 11.59 & 0.48 & \textbf{0.34} & \textbf{0.65} & 0.48 & \textbf{0.74}  \\
        SyntheOcc-FE (ours) w/ multi-view cosine similarity & 13.91 & \textbf{0.49} & \textbf{0.34} & 0.64 & \textbf{0.51} & \textbf{0.74}  \\
        \bottomrule
    \end{tabular}
\end{table*}

\begin{table}[t]
    \centering
    \setlength{\tabcolsep}{3pt}
    \renewcommand{\arraystretch}{0.92}
\caption{\textbf{Downstream 2D Object Detection.} $\mathrm{mAP}$ on the KITTI-360 test set for models trained on real KITTI-360 versus generated training data. Best in bold.}
    \label{tab:downstream_task}
    \begin{tabular}{lcccc}
        \toprule
        \textbf{Training data} & \textbf{Vehicle} $\uparrow$ & \textbf{Person} $\uparrow$ & \textbf{Bicycle} $\uparrow$ & \textbf{Avg} $\uparrow$ \\
        \midrule
        KITTI-360 dataset     & 27.7 & 10.7  & 0.1 & 12.9 \\
        \midrule
        SyntheOcc-FE  & \textbf{25.6 }& 0.6   & \textbf{5.5} & \textbf{10.6} \\
        MIVIFI        & 20.7 & \textbf{4.0}   & 1.1 & 8.6  \\
        \bottomrule
    \end{tabular}

\end{table}
\section{Discussion}
\label{sec:discussion}

\paragraph{Regarding Expandability}
Unlike SyntheOcc-FE, which is constrained by the scarcity of fisheye datasets, MIVIFI can be trained on any perspective dataset, as demonstrated here on nuScenes. Extending this training to additional datasets promises further improvements in generalization and image quality. Furthermore, MIVIFI could be trained with single-view images. This flexibility would further enable extension to other sensors and simulations, for example, those given by the RICO benchmark~\cite{neuwirth2025rico}.
\mbox{\noindent\smalltriangleright~}\textit{MIVIFI is extensible with fisheye and non-fisheye datasets.}

\paragraph{Regarding Generalization vs. Specialization}
The results from SyntheOcc-FE and MIVIFI indicate a general trade-off between generalization and specialization. SyntheOcc-FE yields high-quality results due to its specialization in KITTI-360. MIVIFI, on the other hand, is trained on more diverse data and demonstrates the benefits of this generalization in object, time, and weather edits, but lags slightly in visual quality. Joint training induces domain conflicts, as shown in Tables~\ref{tab:results} and \ref{tab:downstream_task}.
\noindent\smalltriangleright \textit{SyntheOcc-FE is more specialized, and MIVIFI is more general.}

\paragraph{Regarding Perspective Images}
In Figure~\ref{fig:erp_version}, it can be seen that the three perspective images do not match perfectly, creating a faulty training signal, which hinders training. A possible approach would be to train on a single image and mask out more area, or to improve the alignment of the overlapping regions. The same issues would arise when training a model to convert perspective to fisheye images. 
\mbox{\noindent\smalltriangleright~}\textit{The ERP approach enables perspective image training but has alignment shortcomings.}

\paragraph{Regarding Sequences}
With semantic MPI conditioning, new trajectories can be simulated from the occupancy map or existing sequences can be adapted. However, in the current setup, the frames would not be consistent. This consistency can be achieved by incorporating conditioning on previous frames. While SyntheOcc-FE and MIVIFI can be extended to include additional conditioning, this is beyond the scope of this work.
\noindent\smalltriangleright \textit{Consistent sequence editing would require additional conditioning through previous frames.}

\paragraph{Regarding Downstream Application}
We demonstrated that synthetic images provide a useful training signal for 2D object detection. Due to the semantic occupancy map, this can be further extended to panoptic segmentation. Although the overall object detection $\mAP$ is low, the results on real and synthetic data are comparable. We suspect that downstream results could be improved by optimizing the training setup, enhancing generation models, or using more data (which is particularly feasible for our methods). 
\noindent\smalltriangleright \textit{The semantic conditioning enables SyntheOcc-FE and MIVIFI for object detection and panoptic segmentation.}

\section{Conclusion}
\label{sec:conclusion_future_work}
In this work, we introduce the problem of multi-view fisheye image generation and present two approaches to address it: SyntheOcc-FE and MIVIFI. SyntheOcc-FE is trained solely on fisheye data and achieves slightly better visual quality. MIVIFI combines SyntheOcc-FE with Equirectangular Projection, enabling training on additional multi-view perspective datasets such as nuScenes. We show the benefits of this general approach for scene editing, offering a practical method for extending limited fisheye datasets and generating rare edge cases. Future work will expand the training data, incorporate image conditioning for consistent sequence editing, and place greater emphasis on downstream tasks. Overall, our cross-domain strategy reduces the need for costly 3D simulation and helps bridge the gap between scarce fisheye data and the diverse conditions required for training robust 360° perception models.
\clearpage

\begin{center}
    \normalfont\normalsize\textbf{Appendix}
\end{center}
\vspace{0em} 

\setcounter{section}{0}
\renewcommand{\thesection}{\Alph{section}}

\makeatletter
\renewcommand\section{\@startsection {section}{1}{\z@}%
                                   {-1.5ex \@plus -0.5ex \@minus -.2ex}%
                                   {1.0ex \@plus .2ex}%
                                   {\normalfont\normalsize\itshape}}

\renewcommand{\@seccntformat}[1]{%
  \ifnum\pdfstrcmp{#1}{section}=0
    Appendix \csname the#1\endcsname.\hspace{0.5em}%
  \else
    \csname the#1\endcsname\quad
  \fi
}
\makeatother

\setcounter{figure}{0}
\setcounter{table}{0}
\renewcommand{\thefigure}{A.\arabic{figure}}
\renewcommand{\thetable}{A.\arabic{table}}

\section{Single-View ERP}
\begin{figure}[H]
    \centering
    \includegraphics[width=0.5\linewidth]{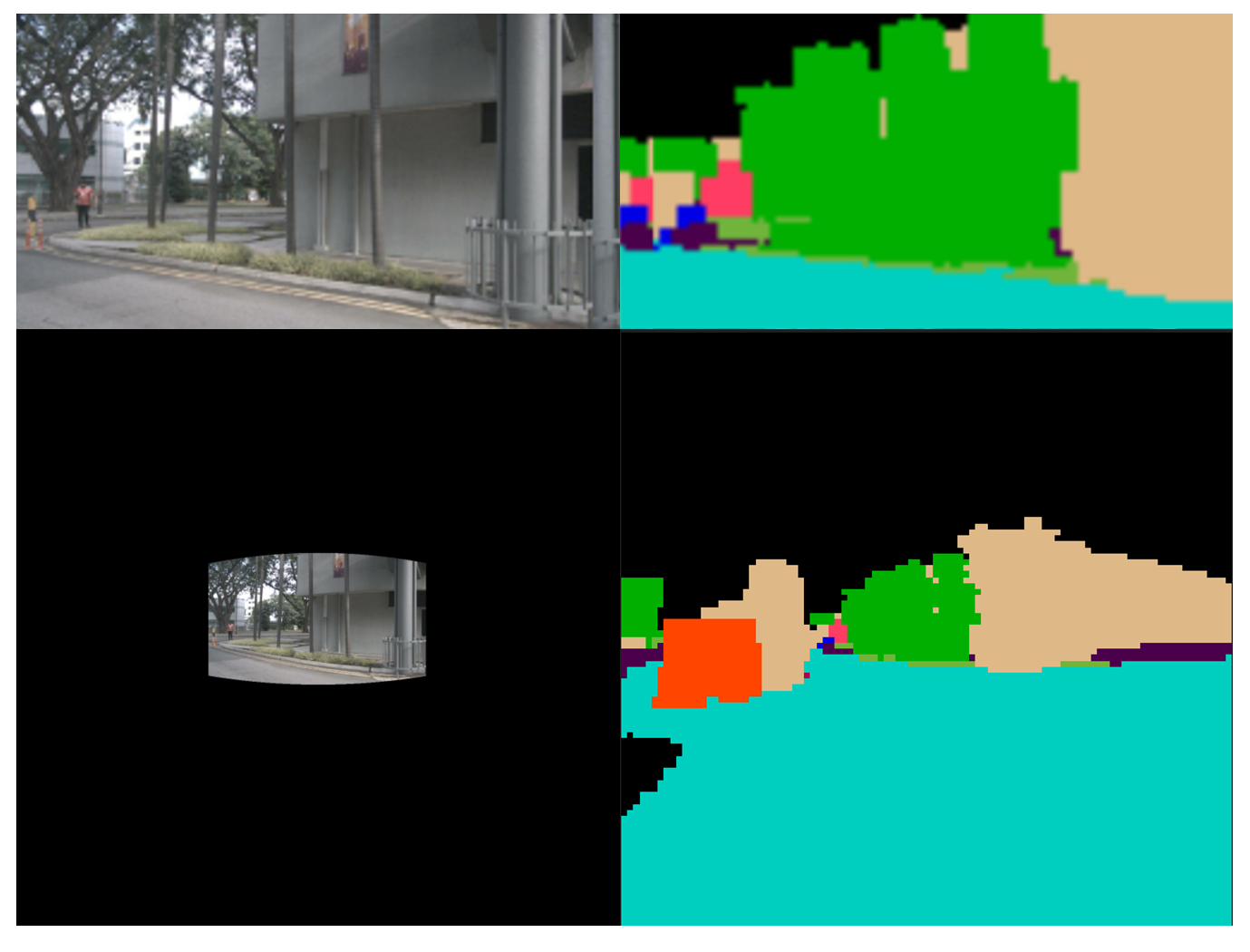}
\caption{\textbf{Single-View ERP visualization.} Top: perspective image and corresponding MPI. Bottom: single-view mapped to ERP with masked regions and full-screen MPI.}
    \label{fig:nuscenes_erp_convesion2}
\end{figure}

MIVIFI can train on single-view images with a full semantic occupancy map and does not require multi-view perspective datasets. Figure~\ref{fig:nuscenes_erp_convesion2} shows a single-view ERP example. This enables single-view training and Out-of-FoV completion.

\section{Adapting Sequenzes}
\begin{figure}[H]
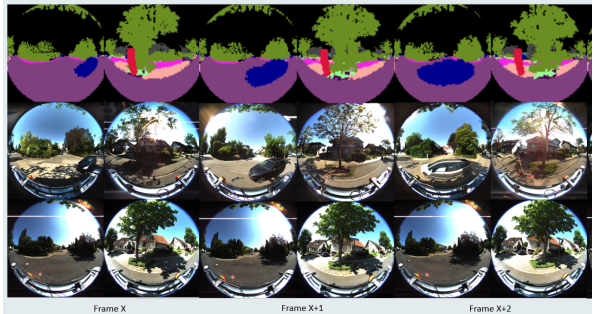

    \centering
    \adjincludegraphics[
        width=0.9\linewidth,
        trim={0 0 {0.4\width} 0},
        clip
    ]{figures/geo_modify_temporal.png}
\caption{\textbf{SyntheOcc-FE object insertion.} Top: occupancy map with added car (blue) and person (red). Middle: generated multi-view fisheye images. Bottom: real scene. The car renders, the person is omitted due to limited training data, and results vary without adjacent-frame conditioning.}
    \label{fig:SyntheOcc-FE-add-new-objects_sequence2}
\end{figure}

Figure~\ref{fig:SyntheOcc-FE-add-new-objects_sequence2} shows sequence adaptation with SyntheOcc-FE. It generates high-quality images and inserts the car correctly, but struggles with the person due to limited data. Without visual conditioning, the images are not consistent.

\section{Qualitative Object Detection Results}
\begin{figure}[H]
    \centering
    \adjincludegraphics[
        width=0.5\linewidth,
        trim={0 0 0 {0.5\width}},
        clip
    ]{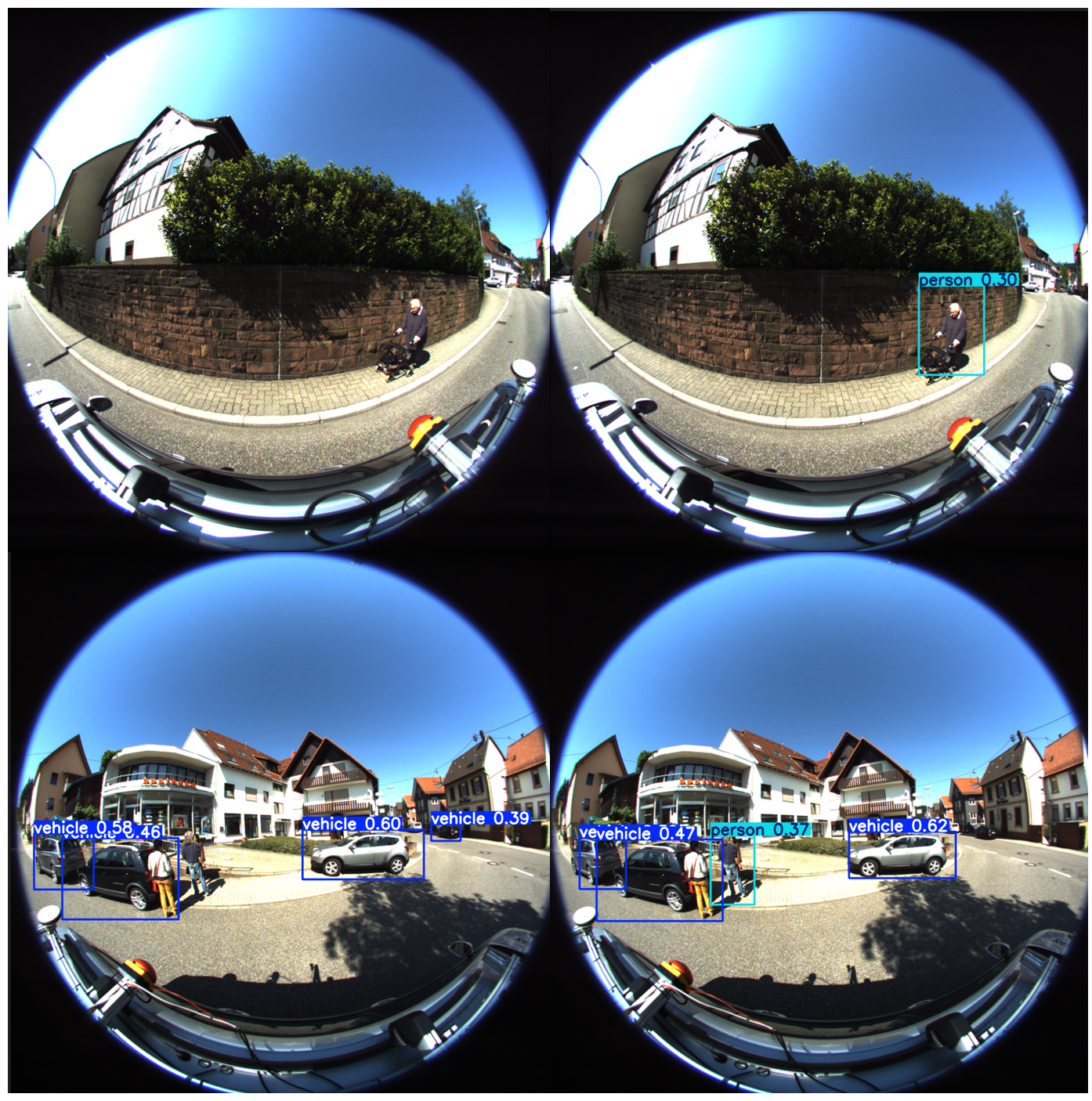}
\caption{\textbf{Qualitative Object Detection Results.} SyntheOcc-FE (left) struggles more with people than MIVIFI (right).}
    \label{fig:yolo_quali}
\end{figure}

Figure~\ref{fig:yolo_quali} shows qualitative results for SyntheOcc-FE and MIVIFI for the downstream object detection task. SyntheOcc-FE detects no pedestrians.

\newpage

\bibliographystyle{splncs04}
\bibliography{bibliography}

\begin{thebibliography}{10}
\providecommand{\url}[1]{\texttt{#1}}
\providecommand{\urlprefix}{URL }
\providecommand{\doi}[1]{https://doi.org/#1}

\bibitem{nuscenes}
Caesar, H., Bankiti, V., Lang, A.H., Vora, S., Liong, V.E., Xu, Q., Krishnan, A., Pan, Y., Baldan, G., Beijbom, O.: nuscenes: A multimodal dataset for autonomous driving. In: CVPR (2020)

\bibitem{magicdrive}
Gao, R., Chen, K., Xie, E., Hong, L., Li, Z., Yeung, D.Y., Xu, Q.: {MagicDrive}: Street view generation with diverse 3d geometry control. In: ICLR (2024)

\bibitem{kitti}
Geiger, A., Lenz, P., Stiller, C., Urtasun, R.: Vision meets robotics: The {KITTI} dataset. IJRR  (2013)

\bibitem{GANs}
Goodfellow, I.J., Pouget-Abadie, J., Mirza, M., Xu, B., Warde-Farley, D., Ozair, S., Courville, A., Bengio, Y.: Generative adversarial nets. NeurIPS  (2014)

\bibitem{depthanycamera}
Guo, Y., Garg, S., Miangoleh, S.M.H., Huang, X., Ren, L.: Depth any camera: Zero-shot metric depth estimation from any camera. In: CVPR (2025)

\bibitem{fid}
Heusel, M., Ramsauer, H., Unterthiner, T., Nessler, B., Hochreiter, S.: Gans trained by a two time-scale update rule converge to a local nash equilibrium. NeurIPS  (2017)

\bibitem{DDPM}
Ho, J., Jain, A., Abbeel, P.: Denoising diffusion probabilistic models. In: NeurIPS (2020)

\bibitem{jia2022visual}
Jia, M., Tang, L., Chen, B.C., Cardie, C., Belongie, S., Hariharan, B., Lim, S.N.: Visual prompt tuning. In: ECCV (2022)

\bibitem{yolo11_ultralytics}
Jocher, G., Qiu, J.: Ultralytics yolo11 (2024)

\bibitem{KG25}
Kim, S., Go, K.: {Edge-case Synthesis for Fisheye Object Detection: A Data-centric Perspective}. arXiv preprint arXiv:2507.16254  (2025)

\bibitem{kumar2023surround}
Kumar, V.R., Eising, C., Witt, C., Yogamani, S.K.: Surround-view fisheye camera perception for automated driving: Overview, survey \& challenges. IEEE Trans. Intell. Transp. Syst.  \textbf{24}(4),  3638--3659 (2023)

\bibitem{syntheocc}
Li, L., Qiu, W., Cai, Y., Yan, X., Lian, Q., Liu, B., Chen, Y.C.: Syntheocc: Synthesize geometric-controlled street view images through 3d semantic mpis. arXiv preprint arXiv:2410.00337  (2024)

\bibitem{kitti360}
Liao, Y., Xie, J., Geiger, A.: {KITTI}-360: A novel dataset and benchmarks for urban scene understanding in 2d and 3d. PAMI  (2022)

\bibitem{fleet_microsoft_2014}
Lin, T.Y., Maire, M., Belongie, S., Hays, J., Perona, P., Ramanan, D., Doll{\'a}r, P., Zitnick, C.L.: Microsoft {{COCO}}: {{Common Objects}} in {{Context}}. In: Eccv (2014)

\bibitem{wovogen}
Lu, J., Huang, Z., Zhang, J., Yang, Z., Zhang, L.: Wovogen: World volume-aware diffusion for controllable multi-camera driving scene generation. In: ECCV (2024)

\bibitem{mei}
Mei, C., Rives, P.: Single view point omnidirectional camera calibration from planar grids. In: ICRA (2007)

\bibitem{neuwirth2025incremental}
Neuwirth-Trapp, M., Bieshaar, M., Paudel, D.P., Van~Gool, L.: Incremental object detection with prompt-based methods. In: ICCVW (2025)

\bibitem{neuwirth2025rico}
Neuwirth-Trapp, M., Bieshaar, M., Paudel, D.P., Van~Gool, L.: Rico: Two realistic benchmarks and an in-depth analysis for incremental learning in object detection. In: ICCVW (2025)

\bibitem{SD}
Rombach, R., Blattmann, A., Lorenz, D., Esser, P., Ommer, B.: High-resolution image synthesis with latent diffusion models. In: CVPR (2022)

\bibitem{bevgen}
Swerdlow, A., Xu, R., Zhou, B.: Street-view image generation from a bird's-eye view layout. Robotics and Automation Letters  (2024)

\bibitem{curveddiffusion}
Voynov, A., Hertz, A., Arar, M., Fruchter, S., Cohen-Or, D.: Curved diffusion: A generative model with optical geometry control. In: ECCV (2024)

\bibitem{ssim}
Wang, Z., Bovik, A., Sheikh, H., Simoncelli, E.: Image quality assessment: from error visibility to structural similarity. IEEE Transactions on Image Processing  \textbf{13}(4),  600--612 (2004). \doi{10.1109/TIP.2003.819861}

\bibitem{surroundocc}
Wei, Y., Zhao, L., Zheng, W., Zhu, Z., Zhou, J., Lu, J.: Surroundocc: Multi-camera 3d occupancy prediction for autonomous driving. arXiv preprint arXiv:2303.09551  (2023)

\bibitem{bevcontrol}
Yang, K., Ma, E., Peng, J., Guo, Q., Lin, D., Yu, K.: Bevcontrol: Accurately controlling street-view elements with multi-perspective consistency via bev sketch layout. arXiv preprint arXiv:2308.01661  (2023)

\bibitem{controlnet}
Zhang, L., Rao, A., Agrawala, M.: Adding conditional control to text-to-image diffusion models. In: ICCV (2023)

\bibitem{lpips}
Zhang, R., Isola, P., Efros, A.A., Shechtman, E., Wang, O.: The unreasonable effectiveness of deep features as a perceptual metric. In: CVPR (2018)

\bibitem{unipc}
Zhao, W., Bai, L., Rao, Y., Zhou, J., Lu, J.: Unipc: A unified predictor-corrector framework for fast sampling of diffusion models. NeurIPS  (2023)

\end{thebibliography}

\end{document}